\documentclass[sigconf,nonacm]{acmart}
\usepackage{amsmath}
\usepackage{natbib} 
\usepackage{mathtools}
\usepackage[toc,page, title, titletoc]{appendix}
\usepackage{pgfplots}
\pgfplotsset{compat=newest}
\pgfplotsset{scaled y ticks=false}
\usepgfplotslibrary{groupplots}
\usepgfplotslibrary{dateplot}

\usepackage{tikz}
\usepackage{multirow}

\AtBeginDocument{%
  \providecommand\BibTeX{{%
    \normalfont B\kern-0.5em{\scshape i\kern-0.25em b}\kern-0.8em\TeX}}}
    
\begin{document}

\title[]{Spatiotemporal downscaling and nowcasting of urban land surface temperatures with deep neural networks}

\author[S.Kurchaba]{Solomiia Kurchaba}
\email{skurchaba@tudelft.nl}
\orcid{0000-0002-0202-1898}
\affiliation{
\institution{Department of Geoscience and Remote Sensing}
\institution{Delft University of Technology}
\city{Delft}
\country{The Netherlands}
}
\affiliation{
\institution{School of Engineering and Computer Science, }
\institution{Bern University of Applied Sciences}
\city{Biel}
\country{Switzerland}
}

\author[A. Meyer]{Angela Meyer}
\email{Angela.Meyer@tudelft.nl}
\affiliation{
\institution{Department of Geoscience and Remote Sensing}
\institution{Delft University of Technology}
\city{Delft}
\country{The Netherlands}
}
\affiliation{
\institution{School of Engineering and Computer Science, }
\institution{Bern University of Applied Sciences}
\city{Biel}
\country{Switzerland}
}

\date{May 2026}

\begin{abstract}
 
Land Surface Temperature (LST) is a key variable for various applications, such as urban climate and ecology studies. Yet, existing satellite-derived LST products provide either high spatial or high temporal resolution, resulting in a fundamental trade-off between the two. To address this trade-off, we combine observations from a geostationary and a polar orbiting satellite and provide LST fields at high spatial and high temporal resolution (1 km at 15-min intervals). We demonstrate their application for intraday forecasting of LSTs.
To estimate LST fields at high spatiotemporal resolution, a U-Net model is trained to map LST fields from SEVIRI/MSG (3 km and 15 min resolution) along with corresponding solar zenith angles to LST fields from Terra/Aqua MODIS (1 km, 4 overpasses per day) that are collocated in space and time. 
The presented model has been trained on LSTs across large European cities with a population exceeding 1 million inhabitants, and 
achieves an RMSE = $1.92$~°C and near-zero bias MBE = $0.01$~°C on the hold-out test set. As a second step, we present an LST nowcasting model based on ConvLSTM architecture, trained across downscaled LST fields with forecast lead times of 15 to 75 minutes. The nowcasting model outperforms a persistence and a Climatological Rolling Median benchmarks, with RMSEs of $0.57$ to $1.15$~°C for the considered lead times and biases ranging from $-0.1$ to $0.14$~°C. An additional validation conducted against independent MODIS overpasses confirms robust performance.  
Our LST forecast model at high spatiotemporal resolution is directly applicable to operational satellite-based LST monitoring.

\end{abstract}
\maketitle

\section{Introduction}
\label{sec:introduction}
Land surface temperature (LST) is a critical parameter in various fields. It plays an important role in the monitoring of climate change, urban heat islands, droughts, and heatwaves, being also relevant in such fields as agriculture, hydrology, weather forecasting, ecosystem monitoring, and more generally in surface energy balance studies.
As we are facing unprecedented challenges related to global warming, one of the most relevant applications of satellite-derived LSTs is the detection and characterization of surface urban heat islands. The impact of urban heat islands is particularly critical in rapidly growing cities worldwide, where urbanization, land use changes, and population density amplify thermal stress and energy demand \cite{yeboah2025influence}. Since LST can vary substantially over short time and small spatial scales, an accurate short-term forecasting of high-resolution LSTs becomes crucially important for understanding and management of urban heat islands \cite{voogt2003thermal}, as well as a wide range of other applications such as ecosystem monitoring \cite{karnieli2010use} and energy demand forecasting \cite{mirasgedis2006models}. 

Despite their high potential, several challenges must be addressed to enable the effective use of satellite-derived LSTs in urban applications.
A major limiting factor is the trade-off between spatial and temporal resolution of the available satellite-derived LST field estimates.
Geostationary sensors such as Meteosat Spinning Enhanced Visible and Infrared Imager (SEVIRI) usually provide observations with high temporal resolution of several observations per hour, while having lower spatial resolution of around 3 km at nadir in mid latitudes. 
Sensors in low Earth orbit offer a higher spatial resolution of 1 km or less, but typically at most two observations per day \cite{shi2021urban}, which is insufficient for studying sub-diurnal LST processes \cite{stewart2021time, parlow2021regarding, shi2021urban}.

Several studies have investigated the problem of downscaling LST from geostationary satellites across urban areas (e.g. \cite{zakvsek2012downscaling, bechtel2012downscaling, bah2022spatial}) using various statistical methods.
For example, in \cite{keramitsoglou2013downscaling} the authors used several support vector machine (SVM) models to downscale geostationary LST to a 1~km resolution for Athens. 
In \cite{weng2014modeling}, a least square support vector machine was used to downscale geostationary LST to 1~km resolution, for one day per season during 2012. 
In \cite{hurduc2024multi}, the authors used a multi-layer perceptron model to downscale LST from approximately 4.5~km to 750~m for the city of Madrid. 
However, all the above-mentioned studies used classical machine learning models that use human-defined features and require spatial integration of the data, which usually results in the loss of information and sub-optimal model performance.
In \cite{chang2024generating}, the authors used a convolutional neural network (CNN) to downscale LST fields from 2 km to 70 meters for the city of Los Angeles. 
The study uses LST from the GOES-R geostationary sensor as input, and LST from ECOSTRESS as a target variable. 
However, an average revisiting time of an ECOSTRESS satellite is 3-4 days, which makes it difficult to collect enough data for the training of a deep learning model for multiple urban areas, especially when considering urban areas with a limited number of days suitable for satellite observations. 
Such a limitation makes the proposed approach difficult to scale.

Research on forecasting of LST fields at intraday timescales remains limited, particularly for products derived from Meteosat satellites or across European cities. 
In \cite{kartal2022prediction}, several machine learning models were proposed for one-step-ahead forecasting of MODIS LST. While maintaining the spatial resolution of 1~km, such a model operates at the native time resolution of MODIS, and therefore cannot provide intraday forecasts. In \cite{brown2024toward}, the authors introduced a deep learning architecture for nowcasting very high resolution (30~m) LST. 
This approach relies on Landsat-8 measurements to achieve the fine spatial detail, but as a result, the forecasts are tied to a single typical overpass time (approximately 10:30~am local time). Moreover, the 16-day revisit cycle of the satellite limits the method’s ability to support intraday or even medium-range LST forecasting. 

Our study introduces the first intraday satellite-based deep learning forecast model of high-resolution LST fields. We also demonstrate the first deep learning LST downscaling model that enables 1 km resolution LSTs updated every 15 minutes across Europe.
We propose a combined LST downscaling and nowcasting approach. First, we develop a pan-European LST downscaling model trained using SEVIRI-derived LST fields as input and co-located MODIS-derived LST fields as high-resolution targets. This model is designed to generalize across major European urban areas (population > 1 million), enabling the generation of high-resolution (1 km), high-frequency (15-minute) LST fields across the continent.

Building on this generalized capability, we then focus on a city-scale application to demonstrate the practical utility of the downscaled data for short-term forecasting. We present an LST nowcast model that forecasts LST fields across European cities for up to 75 minutes ahead. We demonstrate our approach for three representative European cities: Bucharest, Antwerp, and Berlin.
This two-step approach allows us to first establish a robust, transferable spatial downscaling framework and subsequently assess its value in a localized, operational nowcasting setting.
Finally, we evaluate our proposed LST downscaling and nowcasting approach against actual MODIS observations.

The main contributions of this study are as follows:
\begin{itemize}
\item We present the first deep learning LST downscaling model that enables 1 km resolution LSTs updated every 15 minutes across Europe. Our model maps 15-min 3-km SEVIRI LST to 1-km MODIS LST with 2-4 overpasses per day.
 \item Our downscaling model is designed to generalize across many urban areas, not just one, including all major European cities (population$>$1 million) at a low error and a near-zero bias. 
 \item We introduce the first intraday satellite-based deep learning nowcasting model for high-resolution LST fields and demonstrate its application at the city scale using Bucharest, Antwerp, and Berlin as representative case studies. Building on the downscaled 1 km, 15-minute LST fields generated by our pan-European model, this nowcasting framework enables short-term urban LST forecasting with lead times ranging from 15 to 75 minutes. This two-step design highlights the practical value of the downscaling approach by bridging continental-scale data generation with localized forecasting applications. Our proposed nowcasting model significantly outperforms the two data-driven benchmark models and shows strong agreement with MODIS LST observations.
  \end{itemize}

This paper is organized as follows. In Section~\ref{sec:data_process},  we describe the data used in this study, as well as the preprocessing workflow used to construct training and test datasets for the LST downscaling and nowcasting approaches. Section~\ref{sec:method} presents the methodological framework, including the U-Net downscaling model, the ConvLSTM nowcasting method, and the benchmark predictors. In Section~\ref{sec:results}, we report the results of the performed experiments for LST downscaling and nowcasting, followed by a further validation against MODIS observations. Finally, in Section~\ref{sec:discussion}, we discuss the main findings, limitations, and directions for future work.

\section{Data and Preprocessing}
\label{sec:data_process}
\subsection{Data sources}
In this study, we integrate several satellite and auxiliary datasets to construct a multi-sensor machine learning dataset for the downscaling and nowcasting of LST fields.
\subsubsection{SEVIRI/MSG LST}
We use the clear-sky Land Surface Temperature (MLST) product \cite{trigo2009algorithm} provided by the Satellite Applications Facility on Land Surface Analysis (LSA SAF) of the European Organization for the Exploitation of Meteorological Satellites (EUMETSAT). The LSTs were derived from spectral radiance measurements of the Spinning Enhanced Visible and Infrared Imager (SEVIRI) onboard the Meteosat Second Generation (MSG) \cite{schmetz2002introduction} geostationary Earth monitoring satellite series. SEVIRI provides full-disk observations with a viewing zenith angle between 0° and 80°. It offers a temporal resolution of 15 minutes and a spatial resolution at nadir of approximately 3 km (0.05°). 
\subsubsection{MODIS LST}
As a high-resolution LST target variable, we use daily Level-2 LST products \cite{hulley2021modis} MOD21 (Terra) and MYD21 (Aqua), product version: 0.61. These products provide global coverage at 1 km spatial resolution at nadir. Equatorial overpass times are approximately 10:30 a.m./p.m. local solar time for Terra and 1:30 a.m./p.m. for Aqua.

\subsubsection{Auxiliary Data}

We include the Solar Zenith Angle (SZA), computed for each pixel at the satellite overpass time using the pvlib Python package (v0.13.0), as an auxiliary predictor. SZA directly controls the amount of incoming shortwave radiation reaching the surface and is therefore a primary driver of the diurnal and spatial variability of LST. Including SZA introduces physically meaningful information on solar illumination geometry, helping the model account for insolation-driven temperature variations and improving the physical consistency of the predicted LST fields.

We note, however, that SZA represents only a first-order control on radiative forcing. Other factors such as land cover, vegetation state (e.g., NDVI), urban morphology (e.g., sky view factor or building volume), and anthropogenic influences (e.g., population density or pollution) also affect the surface energy balance through mechanisms such as heat storage, radiative trapping, and anthropogenic heat release. These variables are not explicitly included in the current framework. Instead, we assume that part of their aggregated effect is implicitly encoded in the coarse-resolution LST signal used as input.
This design choice allows us to develop a sensor-driven and spatially transferable approach that does not rely on auxiliary datasets, which are often heterogeneous, static, or unavailable at appropriate spatial and temporal resolutions. 

\subsection{Data preprocessing}
We make use of 21 years of satellite observations (2004–2025). Because our primary application is the characterization of urban heat islands \cite{yang2016research}, we restrict the dataset to the climatologically warmest months of the year: 15 May–15 September. Training data are collected for European cities with populations exceeding one million inhabitants, ensuring diversity in climatic, morphological, and land-cover characteristics. As a hold-out test set for both downscaling and nowcasting models, we use the satellite measurements from the years 2007, 2013, 2019, and 2025. The rest of the years from the studied period are used for the training and validation of the models. 

\subsubsection{Downscaling}
To prepare the data for the training of the LST downscaling model, we perform the following data preprocessing steps. First, we acquire MODIS LST Level-2 granules and perform quality filtering based on quality flag values (only pixels with good or nominal quality are kept), LST accuracy (only pixels with good or excellent LST retrieval performance are kept), viewing angles ($<50$°), and cloud flag (cloud, cloud shadow, or cirrus pixels are removed). In addition, we perform the removal of outliers: LST pixels with values below 0~°C and above 65~°C are removed. 

Next, we spatially subset the data to patches covering each target city and regrid these patches onto a common, uniform grid of a resolution $0.01$°. The regridding step is necessary because the original MODIS data are provided on a swath-based grid, where pixel locations and spacing vary with satellite viewing geometry. By “uniform grid,” we refer to a regular, fixed-resolution grid with consistent spacing in latitude and longitude, which ensures spatial alignment across samples. Only images with less than 50\% missing or poor-quality pixels (i.e., pixels removed during filtering) are retained. These images serve as the ground-truth targets for training the LST downscaling model. 

For each MODIS overpass included in the training set, we identify the temporally closest SEVIRI acquisition. Since SEVIRI observations are available every 15 minutes, the time difference between the paired MODIS and SEVIRI acquisitions is always less than 15 minutes. SEVIRI scenes undergo the same spatial subsetting and are additionally filtered to remove cloud- or water-contaminated pixels before being regridded to the same uniform grid. Due to the coarser spatial resolution of SEVIRI, each pixel represents a larger area and thus has a higher likelihood of being affected by clouds or other invalid conditions. To ensure sufficient data quality, we therefore impose a stricter requirement on data coverage compared to MODIS: only scenes with more than 80\% valid pixels (i.e., pixels remaining after filtering) are retained.

\begin{figure*} 
   \centering
    \includegraphics[width=0.9\textwidth]{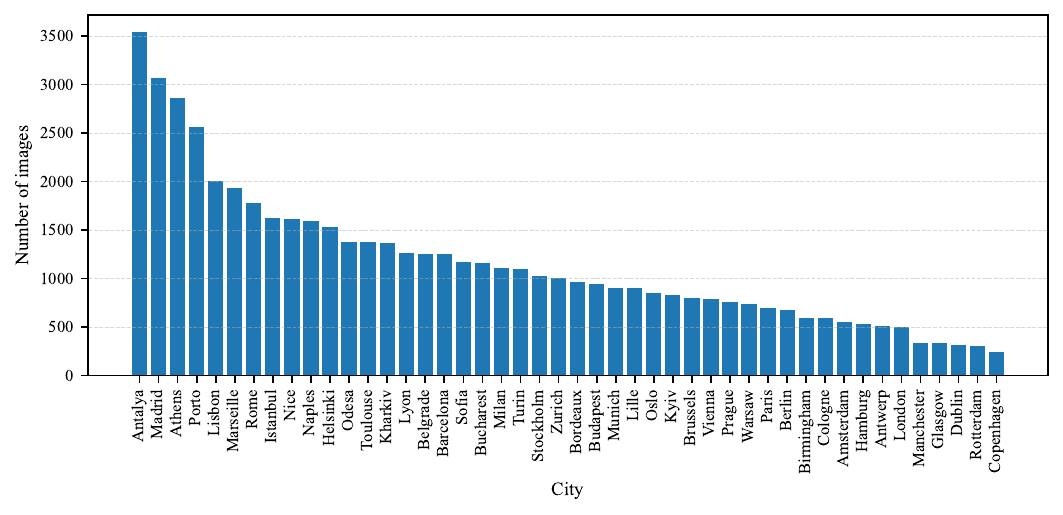}
  \caption{Distribution of image patches per city in the dataset before training-test split and dataset balancing. For model training, a maximum of 1500 samples per city is used to ensure a more uniform representation across regions.}
    \label{fig:images_per_city}
\end{figure*}

We only retain MODIS–SEVIRI pairs when at least 75\% of their valid-pixel areas spatially overlap. All thresholds were derived from an empirical examination of MODIS–SEVIRI pairs to balance the size of the resulting dataset and the image quality. The distribution of image patches per city is shown in Fig. \ref{fig:images_per_city}. To ensure a balanced representation across cities and to prevent over-representation of specific climatic regions, we limit the number of training samples to a maximum of 1500 images per city. This downsampling strategy mitigates biases toward cities with higher data availability and promotes more uniform learning across different geographic and climatic conditions. Ultimately, our dataset contains $52938$ datapoints.
 
\subsubsection{Nowcasting}
 We include all available SEVIRI acquisitions for the studied period of 2004-2025 in the training of the LST nowcasting model. We spatially subset the data around cities of interest, and regrid the resulting image patches to the grid used for the LST downscaling model. We apply the pre-saved downscaling model to estimate the prepared SEVIRI image patches at higher spatial resolution. The resulting patches are then arranged into time-series sequences to be used as an input to the nowcasting model.
 Ultimately, for Bucharest, Antwerp, and Berlin, the training set is composed of respectively $98330$, $54031$, and $51237$ datapoints. The test set contains $24152$, $12042$, and $11834$ datapoints respectively.

\section{Method}
\label{sec:method}

This Section presents the methodological framework of the proposed downscaling--nowcasting pipeline. We first describe the formulation of the LST downscaling task and the U-Net architecture used to reconstruct high-resolution MODIS-like LST fields from lower-resolution SEVIRI inputs. We then introduce the nowcasting setup, including the temporal prediction problem, the construction of training samples, and the ConvLSTM-based model used to forecast future high-resolution LST fields. Finally, we define the benchmark predictors used for evaluation.

The proposed framework follows a two-stage design, where spatial downscaling and temporal prediction are learned separately. This choice is motivated by both methodological and practical considerations. First, the spatial and temporal components address distinct learning problems: the U-Net focuses on reconstructing spatial detail from coarse-resolution observations, while the ConvLSTM captures temporal dynamics of already downscaled fields. Decoupling these tasks simplifies training and allows each model to specialize in a well-defined objective. Moreover, the availability of supervision differs between tasks. High-resolution MODIS LST provides a direct target for training the spatial downscaling model, whereas temporally consistent high-resolution sequences are not available without applying a downscaling model. The sequential design enables the use of all available high-resolution snapshots for spatial learning, while leveraging temporally dense SEVIRI observations for forecasting.

\subsection{Downscaling}
Formally, the LST downscaling task is defined as a map
\begin{equation}
    Y=f(X)
\end{equation}

where $X$ is the low-resolution SEVIRI LST field (augmented as in our case by SZA or potentially other auxiliary predictors), $Y$ is the corresponding high-resolution MODIS LST field, and $f$ is a regression function approximated via a machine learning model. In this work, we deliberately restrict the predictor space to LST fields and solar illumination geometry (SZA), in order to investigate how much of the kilometer-scale spatial variability can be recovered from satellite-derived thermal signals alone. Thus, the proposed framework should be interpreted as a data-minimal and sensor-driven approach, rather than a fully physically explicit urban climate model.

We implement a convolutional encoder–decoder U-Net architecture \cite{ronneberger2015u} (architecture details can be found in Appendix~\ref{app:downscaling_specs}). A U-Net consists of a contracting path that captures contextual information through successive convolution and pooling operations, and an expansive path that progressively reconstructs high-resolution predictions. Skip connections between encoder and decoder layers help preserve fine-scale spatial patterns, making the U-Net particularly effective for tasks requiring both global context and local detail. In remote sensing, U-Net architectures have been successfully applied to cloud segmentation \cite{jiao2020refined}, land-cover and land-use classification \cite{zhang2018urban}, and image super-resolution \cite{xu2018high}, demonstrating robustness under varying spatial resolutions and sensor characteristics.

Because the cities in our dataset vary in spatial extent, input images are padded to a standardized dimension of 128×128 pixels, which is computationally efficient and compatible with the downsampling scheme of the U-Net. 
To fill missing values in the input, we use k-nearest neighbors (KNN) imputation with k = 5, following common practice. We note that, in urban environments, LST fields can exhibit strong spatial heterogeneity with sharp thermal gradients (e.g., between built-up areas, vegetation, and water bodies), which may not be fully captured by the local smoothness assumption underlying KNN. 
As a result, this approach can introduce localized smoothing or interpolation artifacts, particularly when the fraction of missing pixels is high. 
However, in our dataset, missing values typically affect a limited portion of each image, and KNN imputation provided a practical trade-off between simplicity and performance. 
We did not perform explicit hyperparameter optimization for k, as the focus of this study is on the downscaling model rather than the imputation method. 
In addition, such experiments would significantly increase computational costs. 
During preliminary experiments, this method yielded better downstream performance than image-wise mean and median imputation, as reflected by lower RMSE (and consistent trends in other regression metrics) on a validation set. Moreover, any potential smoothing effects introduced during imputation are partially mitigated by the learning-based downscaling model, which is trained to recover high-resolution spatial variability from the input data.

\subsection{Nowcasting}
We refer to intraday forecasts with lead times up to 75 minutes as nowcasting and use these terms interchangeably here. We define a nowcasting task as follows:
\begin{equation}
\hat{X}_{t+\Delta t}=g\left(X{t-2\delta t},X_{t-\delta t},X_t\right)
\end{equation}
where $X_t$ denotes the downscaled SEVIRI LST field at time $t$, $\delta t=15$ min is the native SEVIRI sampling interval, $\Delta t \in {15,30,45,60,75}$ min is the prediction lead time, and $g$ is a non-linear mapping learned by the model. We use the three most recent time steps as input to capture short-term temporal dynamics while keeping the input dimensionality manageable. We did not perform an explicit hyperparameter search over the number of input frames, as preliminary experiments indicated that three time steps provide a reasonable trade-off between performance and computational cost.

To construct the training samples, we use trailing temporal windows of three consecutive downscaled SEVIRI-derived LST fields as predictors, while the target is the future downscaled LST field at lead time $\Delta t$ at MODIS resolution. 
A separate model is trained for each lead time using a corresponding dataset.

For the task of nowcasting, we use a Convolutional LSTM Network, which combines recurrent temporal memory with convolutional operators \cite{shi2015convolutional, app142311315, vukotic2017one}. For architecture specification, see Appendix~\ref{app:nowcast_specs}. Such a model processes spatiotemporal data by stacking ConvLSTM layers to capture spatial features and temporal dependencies simultaneously. We use an encoder-decoder structure as it is shown to be effective for time-dependent images or video predictions \cite{app142311315, vukotic2017one}. This architecture is suitable for LST nowcasting because it preserves spatial coherence (e.g., intra-urban thermal gradients) while learning temporal dependencies in image sequences, required for sub-hourly urban temperature nowcasting.

\subsubsection{Benchmarks}
To evaluate the ConvLSTM-based LST forecast model, we compare it against two complementary data-driven baseline predictors: the Persistence and Climatological Rolling Median benchmarks.

The Persistence benchmark assumes the LST field does not change as time progresses, so the Persistence benchmark propagates the latest available field to all forecast lead times. Formally, for each $\Delta t \in \{15,30,45,60,75\}$ min,
\begin{equation}
    \hat{X}^{\mathrm{persistence}}_{t+\Delta t}=X_t.
\end{equation}
Persistence is a reasonable baseline in high-frequency geophysical nowcasting because LST fields are usually strongly autocorrelated in time at intraday forecast horizons.

The Climatological Rolling Median benchmark accounts for the typical diurnal behavior at each time slot while remaining insensitive to short-term fluctuations. For a given timestamp $t$ and lead time $\Delta t$, the prediction is defined pixel-wise as
\begin{equation}
    \hat{X}^{\mathrm{climatology}}_{t+\Delta t}=\mathrm{median}\left\{X_{t+\Delta t-k\cdot 24\mathrm{h}}\right\}_{k=1}^{5},
\end{equation}
that is, the median over the previous five days at the same local time of day and location.

These two benchmarks are relevant and complementary for LST nowcasting: Persistence tests whether the model improves over a strong short-memory predictor, while Climatological Rolling Median tests whether the model captures event-specific departures from the expected diurnal cycle. Demonstrating gains over both baselines indicates that the model learns both immediate temporal evolution and non-stationary dynamics.

\section{Results}
\label{sec:results}

In this Section, we present the results of our experiments. We start with the evaluation of the LST downscaling model, which will then be used for improving the spatial resolution of the inputs of the LST nowcasting model. Next, we present an evaluation of the performance of the nowcasting model. Finally, we present an evaluation of the whole downscaling-nowcasting pipeline against the actual high-resolution MODIS-derived LST fields at the available timestamps.

\subsection{LST downscaling model}

\begin{table}
    \centering
    \begin{tabular}{cccc}
    \hline
    \textbf{R$^2$} & \textbf{RMSE [°C]} & \textbf{MAE [°C]} & \textbf{MBE [°C]} \\
    \hline
    0.97 & 1.92 & 1.44 & 0.01 \\
    \hline   
    \end{tabular}
    \caption{\textbf{Performance of the LST downscaling model on the hold-out test set.}}
    \label{tab:downscaling_metrics}
\end{table}

\begin{figure}
   \centering
    \includegraphics[width=0.9\linewidth]{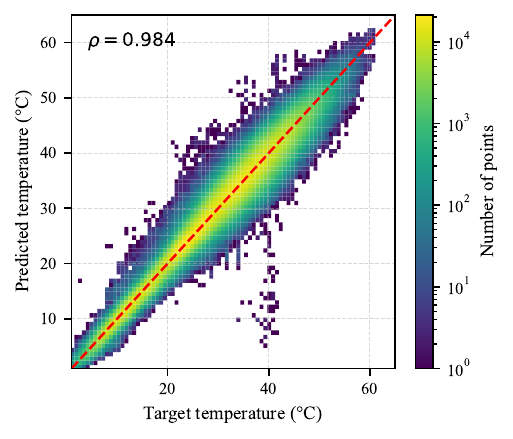}
  \caption{Predictions versus target values of the LST downscaling model. The dashed red line corresponds to 45-degree line. $\rho$ corresponds to the Pearson correlation coefficient.}
    \label{fig:fit_predict_downscale}
\end{figure}

\begin{figure*}
   \centering
    \includegraphics[width=0.9\textwidth]{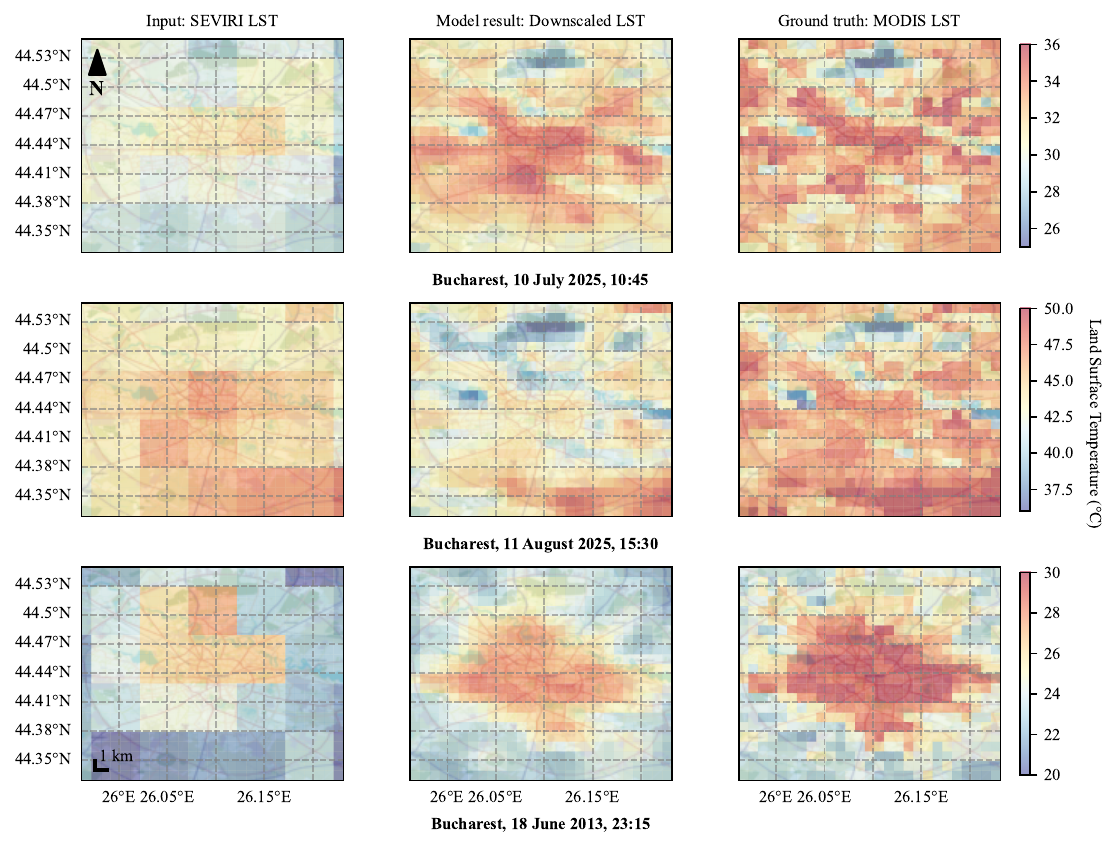}
  \caption{Visual examples of the performance of the downscaling model on the example city of Bucharest. All times are in local summer time of Bucharest (UTC+3). Background maps of Bucharest are generated using python package contextily v.1.6.2 with OpenStreetMap Humanitarian map provider.}
    \label{fig:downscaling_Bucharest}
\end{figure*}

\begin{figure*} 
   \centering
    \includegraphics[width=0.9\textwidth]{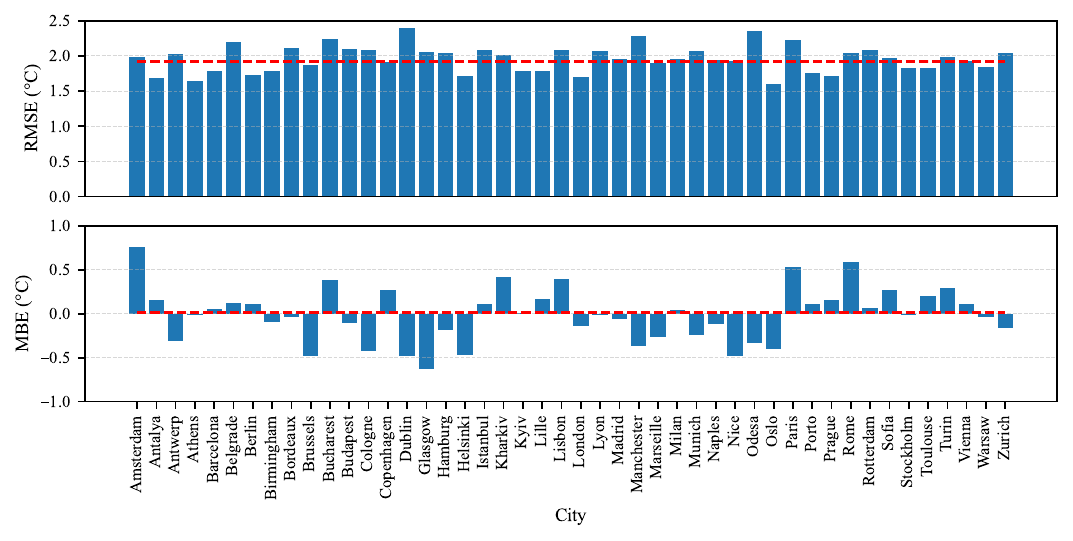}
  \caption{Downscaling performance per city. Top panel: Root Mean Square Error (RMSE) per studied city. Bottom panel: Mean Bias Error (MBE) per studied city. Red dashed lines: RMSE/MBE over the whole dataset.}
    \label{fig:downscaling_city}
\end{figure*}

\begin{figure} 
   \centering
    \includegraphics[width=1.0\linewidth]{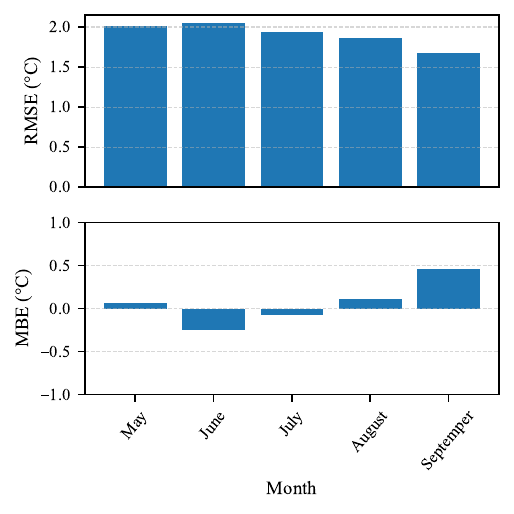}
  \caption{Downscaling performance per month. Top panel: Root Mean Square Error (RMSE) per analyzed month. Bottom panel: Mean Bias Error (MBE) per analyzed month.}
    \label{fig:downscaling_month}
\end{figure}

\begin{figure*} 
   \centering
    \includegraphics[width=0.9\textwidth]{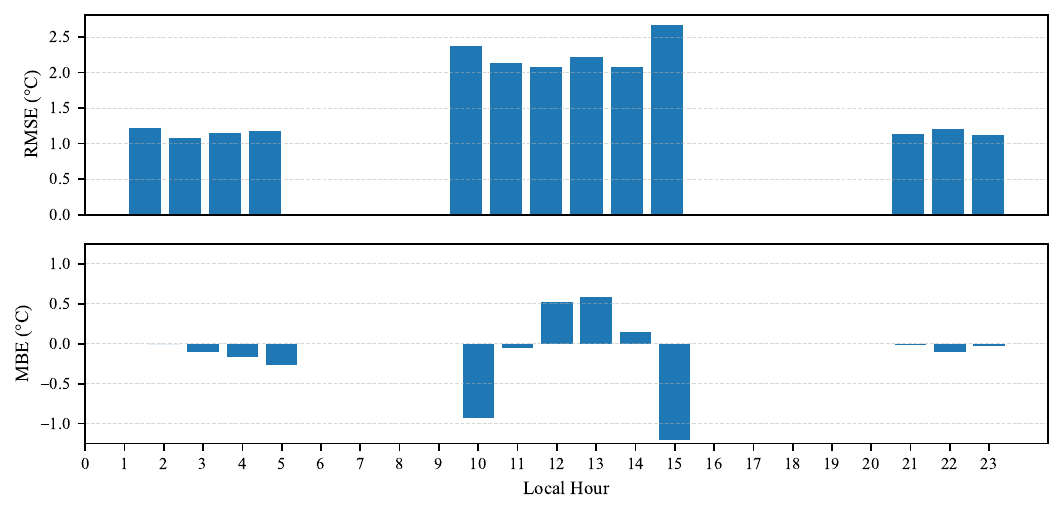}
  \caption{Downscaling performance per hour of day. Top panel: Root Mean Square Error (RMSE) per hour of the day. Bottom panel: Mean Bias Error (MBE) per hour of the day.}
    \label{fig:downscaling_hour}
\end{figure*}

In this Subsection, we present the results of the LST downscaling model. 
The model is trained to map low-resolution SEVIRI LST fields to high-resolution MODIS (Terra/Aqua) LST fields using the U-Net encoder--decoder architecture described in Section~\ref{sec:method}. Evaluation on the hold-out test set shows a strong agreement between predictions and targets (Fig.~\ref{fig:fit_predict_downscale}) with an overall $R^2$ of $0.97$, Root Mean Square Error (RMSE) of $1.92$~°C, Mean Absolute Error (MAE) of $1.44$~°C, and negligible Mean Bias Error (MBE) of $0.01$~°C (see Table~\ref{tab:downscaling_metrics}). The near-zero bias indicates that the model does not systematically over- or underestimate LST across the test set. In Fig. \ref{fig:downscaling_Bucharest}, we provide visual examples of the performance of the downscaling model on the example city of Bucharest over different observation times.

As a next step, we evaluate the robustness of the model across different geographic and temporal conditions by assessing its error metrics by city, month, and hour of day. Fig.~\ref{fig:downscaling_city} summarizes the RMSE and MBE per city included in this study. The spread in RMSE shows that performance is not uniform across all urban areas, reflecting differences in local climate and land-cover heterogeneity.
The city-wise MBE remains close to zero for most cities, suggesting that the model does not introduce strong location-specific biases.

Seasonal variations are shown in Fig.~\ref{fig:downscaling_month}. Errors change across the warm-season months only slightly. The observed changes may reflect changing surface conditions (e.g., vegetation state and soil moisture) and the varying availability of clear-sky observations used to form the training data pairs. 

Finally, Fig.~\ref{fig:downscaling_hour} shows the LST downscaling model performance as a function of local hour. Results cluster around the nominal overpass times of the MODIS Terra and MODIS Aqua instruments, with a spread driven by longitudinal differences across cities, the use of civil time (including daylight saving time) rather than local solar time, and smaller timing offsets introduced by the wide MODIS swath. The hourly patterns indicate that the RMSE of the LST downscaling model depends on the diurnal cycle, just like the LSTs themselves. We observe a noticeably better performance of the model during the night when LSTs are lower. We also notice a deterioration of the model performance for hours that are further away from the typical overpass time of the satellite - this is likely caused by lower training and test data availability for those time stamps.
Despite this variability, the MBE remains small across hours, confirming stable bias behavior throughout the day.

Overall, the results demonstrate that the proposed U-Net-based model can reliably enhance the spatial resolution of 15-min LST fields derived from geostationary satellite while maintaining low bias, which is a key prerequisite for the subsequent nowcasting experiments based on the downscaled LST field sequences.

\subsection{LST nowcasting model}

In this Subsection, we present the results of the nowcasting models. For the nowcasting part of the experiment, we demonstrate the LST nowcasting model for three exemplary cities: Bucharest, Antwerp, and Berlin, selected to represent different urban and climatic conditions across Europe. Bucharest is characterized by a humid subtropical climate (Cfa in the Köppen classification \cite{koppen1884warmezonen}), Antwerp by a temperate oceanic climate (Cfb), and Berlin by a transitional climate between oceanic and continental regimes (Cfb/Dfb).

\begin{figure*}
   \centering
    \includegraphics[width=0.9\textwidth]{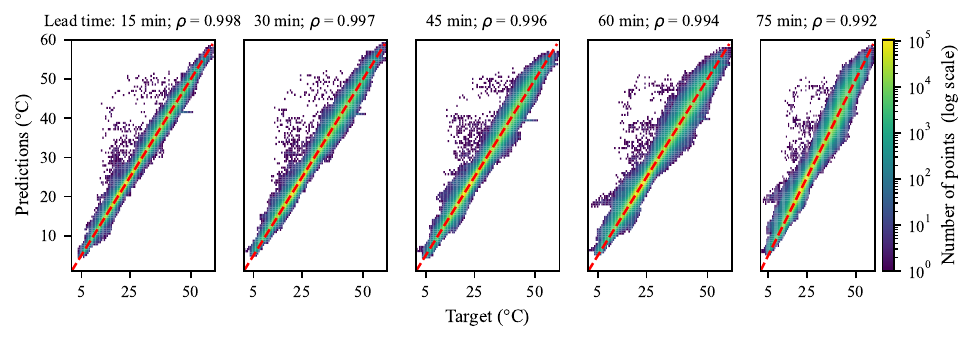}
  \caption{Predictions versus target values of the LST nowcasting model for different lead times for the city of Bucharest. The red line corresponds to 45-degree line. $\rho$ corresponds to the Pearson correlation coefficient.}
    \label{fig:fit_predict_nowcasting}
\end{figure*}

For the nowcasting task, a separate ConvLSTM model was trained for each city, for each lead time using downscaled sequences of SEVIRI-derived LST fields. The performance of the models is evaluated on a hold-out test set.
The aggregated metrics for all three cities are summarized in Table \ref{tab:nowcasting_metrics}. Across all cities, the nowcasting models demonstrate consistently high predictive skill. The $R^2$ values remain very high (0.98–0.99) for all lead times, indicating that the models effectively capture the short-term spatio-temporal evolution of LST. A gradual degradation of performance with increasing lead time is observed, as expected for forecasting tasks. For instance, RMSE increases from approximately 0.57–0.59~°C at 15 minutes to 0.97–1.15~°C at 75 minutes, while MAE increases from about 0.40–0.42~°C to 0.68–0.82~°C. Biases remain small for all cities and lead times, with MBE values close to zero (typically within ±0.1~°C). No systematic over- or underestimation is observed across the forecasting horizon. While small positive or negative biases appear at specific lead times, their magnitude remains negligible compared to the total prediction error. To provide a visual context, Fig. \ref{fig:fit_predict_nowcasting} shows prediction-versus-target scatter plots for Bucharest across different lead times, illustrating a close alignment between predictions and observations. Equivalent plots for Antwerp and Berlin are provided in Appendix~\ref{app:nowcastplotsAntwerpBerlin}.

Despite the common trend, some inter-city differences emerge. From Table \ref{tab:nowcasting_metrics}, we can see that Antwerp and Berlin generally exhibit slightly lower errors than Bucharest, particularly at longer lead times, whereas Bucharest exhibits the strongest increase in error with lead time. 

\begin{table*}
    \centering
    \begin{tabular}{cccccc}
    \hline
   \textbf{City} & \textbf{Lead time [min]} & \textbf{R$^2$} & \textbf{RMSE [°C]} & \textbf{MAE [°C]} & \textbf{MBE [°C]} \\
    \hline
   Bucharest & 15 & 0.99 & 0.59 & 0.42 & 0.04 \\
             &  30 & 0.99 & 0.76 & 0.55 & -0.09 \\
             &  45 & 0.99 & 0.87 & 0.62 & -0.10 \\
             &  60 & 0.99 & 0.99 & 0.70 & -0.04 \\
             &  75 & 0.98 & 1.15 & 0.82 & -0.02 \\
    \hline
    Antwerp & 15 & 0.99 & 0.57 & 0.40 & 0.06 \\
             &  30 & 0.99 & 0.67 & 0.47 & -0.08 \\
             &  45 & 0.99 & 0.77 & 0.55 & 0.01\\
             &  60 & 0.99 & 0.85 & 0.60 & 0.10 \\
             &  75 & 0.98 & 0.97 & 0.68 & 0.04 \\
    \hline    
    Berlin   & 15  & 0.99 & 0.58 & 0.41 & 0.13 \\
             &  30 & 0.99 & 0.67 & 0.48 & 0.13 \\
             &  45 & 0.99 & 0.75 & 0.54 & 0.08 \\
             &  60 & 0.99 & 0.86 & 0.62 &-0.01 \\
             &  75 & 0.99 & 0.97 & 0.70 & 0.04 \\
    \hline   
    \end{tabular}
    \caption{Performance of the LST nowcasting models on the hold-out test set for different lead times.}
    \label{tab:nowcasting_metrics}
\end{table*}

\begin{figure}
 \centering
 \includegraphics[width=1.0\linewidth] {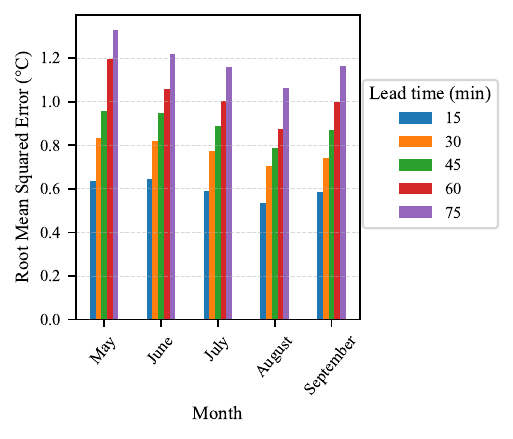}
 \caption{Nowcasting performance per studied month for the city of Bucharest: RMSE per month per lead time.}
 \label{fig:nowcast_month_leadtime_rmse}
\end{figure}

\begin{figure}
 \centering
 \includegraphics[width=1.0\linewidth] {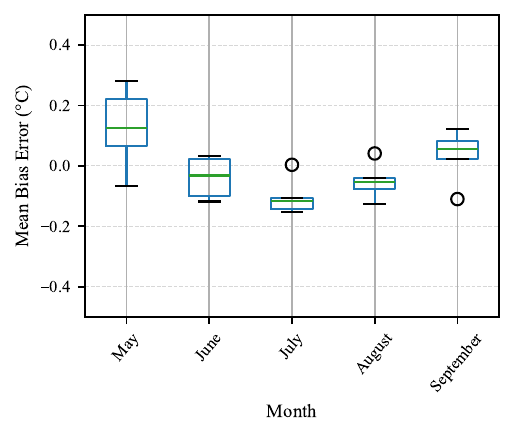}
 \caption{Nowcasting performance per studied month for the city of Bucharest: MBE distribution over lead times per month.}
 \label{fig:nowcast_month_leadtime_mbe}
\end{figure}

We now further break down the results for the city of Bucharest. First, Fig.~\ref{fig:nowcast_month_leadtime_rmse} and Fig.~\ref{fig:nowcast_month_leadtime_mbe} show the distribution of RMSE and MBE per month. Similarly to the downscaling model, we do not see strong variations in the performances of the models across the studied months. Importantly, the relative ranking across lead times is preserved month by month (Fig.~\ref{fig:nowcast_month_leadtime_rmse}): shorter lead times consistently produce lower errors than longer lead times. From Fig.~\ref{fig:nowcast_month_leadtime_mbe}, we also see that even though the MBE exhibits slight seasonal dependency, it remains close to zero for all studied months. We also observe similar patterns for Antwerp and Berlin. Corresponding plots are provided in Appendix~\ref{app:nowcastplotsAntwerpBerlin}.

\begin{figure*}
  \centering
  \includegraphics[width=0.9\textwidth]{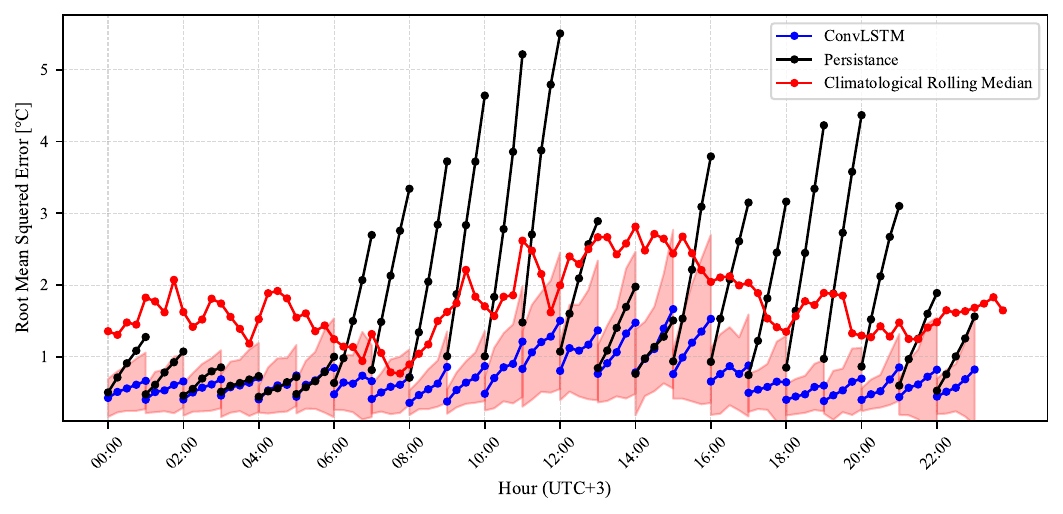}
  \caption{Diurnal evolution of the RMSE per lead time: ConvLSTM-based LST nowcasting model, Persistence benchmark and Climatological Rolling Median benchmark. All times are in local summer time of Bucharest (UTC+3).}
  \label{fig:diurnal_rmse}
\end{figure*}

Finally, with the diurnal analysis in Fig.~\ref{fig:diurnal_rmse} (respectively Fig.~\ref{fig:diurnal_rmse_antwerp}, and Fig.~\ref{fig:diurnal_rmse_berlin} in Appendix~\ref{app:nowcastplotsAntwerpBerlin}), we compare our ConvLSTM approach with two benchmark methods: Persistence and Climatological Rolling Median. Our ConvLSTM-based LST nowcasting models achieve lower RMSEs across most hours of the day and for all lead times, indicating that our LST nowcasting models learn meaningful temporal dynamics beyond simple extrapolation. 
At the same time, the curves reveal two time windows in which outperforming Persistence becomes more challenging, specifically near the transition points of the diurnal cycle where the temperature tendency changes sign. This is visible around approximately 14:00 local time, which is typically the time when LST reaches its daytime maximum and starts to decrease, and between approximately 03:00 and 05:00, when LSTs stabilize following nighttime cooling before increasing again after sunrise. 
Around these points, Persistence can be more difficult to surpass because the recent trajectory is less predictive of the immediate future than during monotonic warming or cooling periods. Taken together, these results show that our proposed LST nowcasting framework provides accurate and stable sub-hourly LST forecasts at 1~km resolution, making it suitable for urban heat monitoring applications.

\subsection{Validation against MODIS-derived LST}

To provide further validation, in this Subsection, we compare the LST forecasts of our model against spatiotemporally collocated MODIS-derived LSTs. In contrast to the previous Subsection, which evaluates the model against downscaled SEVIRI-derived LST fields, here we provide a direct comparison with independent MODIS-derived observations taken from the hold-out test set used for the evaluation of the downscaling model. As in the previous subsection, we consider three cities for the analysis: Bucharest, Antwerp, and Berlin.

\begin{figure*}
  \centering
  \includegraphics[width=0.9\textwidth]{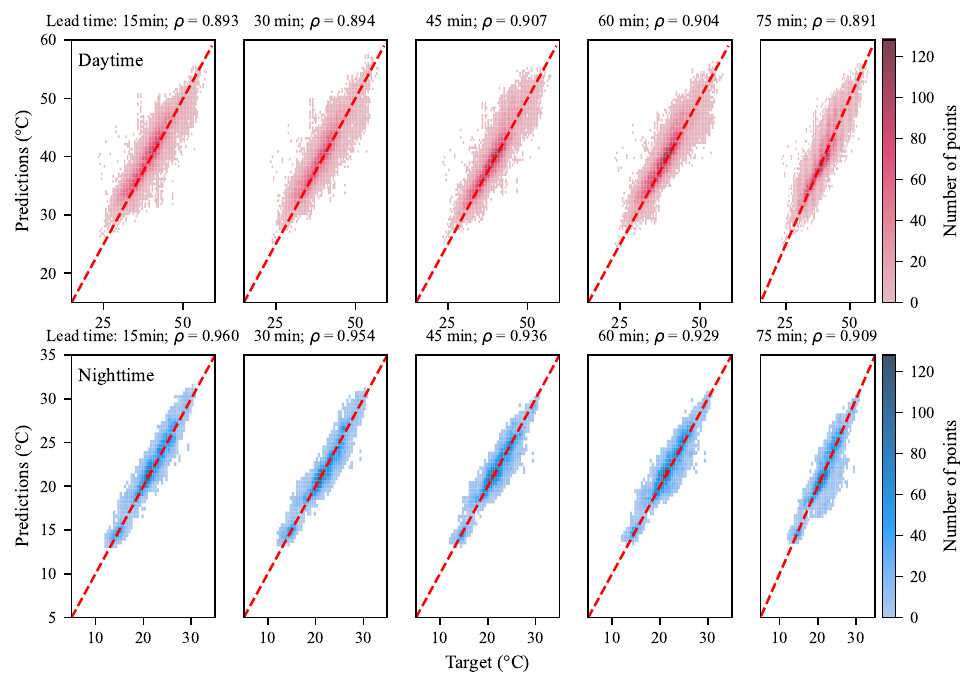}
  \caption{Nowcasting results vs MODIS measurement per lead time for the city of Bucharest. Top panel: LST estimates vs corresponding MODIS-derived LSTs from daytime overpasses of Terra and Aqua. Bottom panel: LST estimates vs corresponding MODIS-derived LSTs from nighttime overpasses of Terra and Aqua. $\rho$ corresponds to the Pearson correlation coefficient.}
  \label{fig:nowcast_vs_modis}
\end{figure*}

We split all available MODIS observations into those corresponding to the daytime and nighttime overpasses of Terra (equatorial overpass times are approximately 10:30 a.m./p.m.) and Aqua (equatorial overpass times are approximately 11:30 a.m./p.m.). The results are presented in Table~\ref{tab:nowcasting_vs_modis_metrics}. The results show a consistent contrast between nighttime and daytime observations. Nighttime performance remains high, with $R^2$ values typically ranging from 0.82 to 0.97 and RMSE between about 0.88 and 1.44~°C across lead times. In contrast, daytime performance is systematically lower, with $R^2$ values ranging more broadly (0.44–0.89) and RMSE values around 1.96–2.7~°C. This day--night contrast is coherent with earlier results from the downscaling subsection, where lower errors were also found at night, likely due to the higher amplitudes of LST variability during the daytime. Similar conclusions can be drawn from Fig.~\ref{fig:nowcast_vs_modis} (also Fig.~\ref{fig:nowcast_vs_modis_antwerp} and Fig.~\ref{fig:nowcast_vs_modis_berlin} in Appendix ~\ref{app:nowcastplotsAntwerpBerlin}).

\begin{table*}
    \centering
    \begin{tabular}{cccccc}
    \hline
    \textbf{City} & \textbf{Lead time [min]} & \textbf{R$^2$} & \textbf{RMSE [°C]} & \textbf{MAE [°C]} & \textbf{MBE [°C]} \\
    \hline
    Bucharest & 15 & 0.78/0.92 & 2.47/1.10 & 1.91/0.86 & 0.66/0.27 \\
              & 30 & 0.79/0.91 & 2.46/1.12 & 1.89/0.88 & 0.57/0.00 \\
              & 45 & 0.81/0.88 & 2.32/1.19 & 1.75/0.93 & 0.55/0.04 \\
              & 60 & 0.80/0.86 & 2.36/1.28 & 1.81/0.99 & 0.65/0.17 \\
              & 75 & 0.78/0.82 & 2.45/1.44 & 1.89/1.05 & 0.44/-0.15 \\
    \hline 
     Antwerp  & 15 & 0.62/0.97 & 2.32/1.09 & 1.75/0.84 & -0.19/-0.09 \\
              & 30 & 0.61/0.97 & 2.33/1.09  & 1.77/0.86 & -0.66/-0.07 \\
              & 45 & 0.61/0.97 & 2.37/1.01  & 1.82/0.81 & -0.40/-0.15\\
              & 60 & 0.58/0.97 & 2.45/1.00  &  1.90/0.79 & -0.50/0.09\\
              & 75 & 0.44/0.96 & 2.70/1.22  &  2.10/0.99 & -0.76/0.08\\
    \hline 
     Berlin   & 15 & 0.89/0.97 & 1.96/0.98 & 1.54/0.74 & -0.20/0.01\\
              & 30 & 0.89/0.97 & 1.97/0.93 & 1.55/0.71 & -0.23/0.09\\
              & 45 & 0.88/0.96 & 2.11/0.89 & 1.68/0.69 & -0.54/0.06\\
              & 60 & 0.87/0.96 & 2.15/0.88 & 1.67/0.68 & -0.31/0.06\\
              & 75 & 0.86/0.96 & 2.25/0.93 & 1.73/0.72 & -0.07/-0.09\\
     \hline 
    \end{tabular}
    \caption{Forecast LSTs compared to MODIS-derived LSTs, by lead time. We report results for daytime/nighttime overpasses of Terra and Aqua.}
    \label{tab:nowcasting_vs_modis_metrics}
\end{table*}

Some inter-city differences can also be identified. Berlin shows the most consistent performance across both daytime and nighttime conditions, maintaining relatively high $R^2$ values and comparatively low errors. Antwerp exhibits very strong nighttime agreement (RMSE below 1.22~°C) but lower daytime performance, particularly at longer lead times, where  $R^2$ decreases more substantially. Bucharest shows intermediate behavior, with stable nighttime performance but slightly higher daytime errors compared to Berlin.

Biases remain relatively small overall but also exhibit a certain structure. Nighttime MBE values are generally close to zero for all cities, indicating the absence of systematic offsets. During the daytime, however, city-dependent biases emerge. Bucharest tends to show a positive bias (slight overestimation of MODIS LST), while Antwerp and Berlin exhibit negative biases. Moreover, the MBE of Berlin tends to be closer to 0 than the MBE of Antwerp and Bucharest. Those trends show similar patterns to what we observe in Figure \ref{fig:downscaling_city} for the downscaling model. Nevertheless, all biases typically remain below 1~°C in magnitude and are small relative to the overall prediction error.
Summing up, despite the slight daytime offset, this MODIS-based validation confirms that the proposed pipeline enables strong predictive skill and remains robust for all studied lead times for high-resolution urban LST nowcasting.

\section{Discussion}
\label{sec:discussion}

In this study, we developed a comprehensive pipeline that combines downscaling and short-term nowcasting of land surface temperatures in urban areas. 
First, we presented a downscaling model trained to increase the spatial resolution of SEVIRI-derived LST fields using MODIS-derived LST as a target. The downscaling model was developed for large European cities and achieved an overall RMSE of 1.96~°C. A closer analysis of the error distribution revealed differences in downscaling quality across cities and months. We also observed that downscaling is generally more challenging for daytime observations than for nighttime observations. 
Nevertheless, MBE remained close to zero in most cases, suggesting that the model generalizes without introducing strong structural biases. Generalization across different climatic regimes is often a concern in data-driven LST modeling. However, based on the per-city evaluation, we do not observe a systematic dependency of model performance on geographic or climatic conditions within the studied domain. In particular, several northern cities exhibit among the lowest downscaling errors, while higher errors are observed in some warmer regions, suggesting that performance is not trivially linked to climate regime.
This indicates that the proposed model is capable of capturing spatial variability across a range of environmental conditions during the warm season.
Overall, these experiments indicated that the developed LST downscaling model is well-suited for preparing inputs for the follow-up LST nowcasting application.

As a second step, we developed LST nowcasting models trained on SEVIRI LST fields that were first downscaled by the downscaling model. For nowcasting, we focused on three cities of interest, Bucharest, Antwerp, and Berlin. We trained five separate LST nowcasting models per city for lead times ranging from 15 to 75 minutes. 
The general RMSE ranged from 0.57-0.59 to 0.97-1.15~°C, gradually increasing with the lead time. 
For all studied cities, the developed nowcasting models outperformed the two baselines, Persistence and Climatological Rolling Median, indicating that the proposed LST nowcasting models capture both immediate temporal evolution and non-stationary LST dynamics. 

Finally, we compared the obtained nowcasting results with actual MODIS-derived LSTs from daytime and nighttime overpasses of MODIS Terra/Aqua. The obtained overall RMSE ranged between 1.96–2.7~°C for daytime predictions and between 0.88–1.44~°C for nighttime predictions. We also observed that nighttime predictions exhibited negligible bias centered around zero, while for daytime predictions, some low city-dependent biases emerged. The observed bias can be partially correlated with the city-specific bias of the downscaling model. The observed differences between daytime and nighttime performance can be interpreted in the context of urban surface energy balance processes. During daytime, LST variability is strongly driven by solar radiation, shadowing effects, and heterogeneous surface properties (e.g., vegetation, building materials), resulting in sharper spatial gradients that are more difficult to reconstruct and predict. In contrast, nighttime LST fields are governed by heat release from urban materials and reduced radiative forcing, leading to smoother spatial patterns and improved predictability. This behavior is consistent with the characteristic dynamics of the surface urban heat island, which is typically more spatially coherent at night.

It is important to note, however, that the observed error values reflect the combined effect of both the LST downscaling and nowcasting models. Due to the sequential design of the framework, any spatial inaccuracies or biases introduced by the downscaling model are propagated into the temporal prediction stage. As a result, the reported performance does not exclusively quantify the predictive skill of the nowcasting model, but rather the joint behavior of the coupled spatio-temporal pipeline.
Moreover, the evaluation is limited by the temporal sampling of MODIS observations, which are concentrated around the overpass times of Terra and Aqua. Consequently, the assessment does not fully capture model performance throughout the diurnal cycle, particularly during periods without high-resolution reference data.

The training and evaluation of the proposed framework are restricted to the period from mid-May to mid-September, corresponding to the climatologically warmest months of the year. This choice is motivated by our primary application, namely the characterization of surface urban heat islands, which are most pronounced during this period. Nevertheless, this temporal restriction limits the applicability of the model for year-round LST monitoring. In particular, the physical drivers of LST variability during colder seasons differ substantially from those in summer, with increased influence of factors such as reduced solar forcing, altered surface energy balance, and season-dependent land–atmosphere interactions. As a result, the relationships learned by the model during warm months may not generalize to winter conditions \cite{an2025spatiotemporal}.
Extending the framework to a full annual cycle would likely require either season-specific models or the inclusion of additional predictors capturing landscape and environmental controls that become more dominant outside the summer period.

To further improve the presented approach, we propose to investigate the influence of the auxiliary data on the obtained results. Urban LST is known to be influenced by such factors as land cover composition, vegetation activity (e.g., NDVI), urban geometry (e.g., sky view factor), and anthropogenic factors (e.g. building density, population activity, and pollution levels). These variables affect surface energy balance components, including heat storage, radiative trapping, and anthropogenic heat release, and therefore contribute to the spatial variability of LST. By not explicitly incorporating these variables, we assume that their effects are encoded in the coarse-resolution LST signal and can be statistically inferred during downscaling. While this assumption enables a simplified and broadly applicable framework, it may limit the physical interpretability of the model and its ability to generalize across heterogeneous urban environments or extreme conditions. While potentially improving spatial detail, such variables may also introduce city-specific dependencies or diurnal biases (e.g., the NDVI–LST relationship is typically stronger during daytime than nighttime \cite{ayanlade2016seasonality,marzban2018influence}), which could reduce the robustness and transferability of the model across regions and times. For the downscaling model, the addition of the variables can be done as additional channels or in a multi-modal fashion \cite{hong2020more}. For the nowcasting model, additional features will have to be split into stationary and dynamic, which, given the used ConvLSTM architecture, could add additional implementation complexity. 

In conclusion, the proposed framework demonstrates that combining deep-learning-based downscaling with short-term nowcasting is a feasible and effective strategy for the diurnal monitoring of urban LST. The results indicate strong predictive performance, low systematic bias of the proposed models, and clear added value over the benchmark predictors. The proposed pipeline can support a broad range of applications, such as near-real-time urban heat island monitoring and forecasting, early warning for short-term heat stress episodes, improved city-scale heat-risk mapping, and targeted cooling interventions. 
It can also provide high-frequency thermal inputs for downstream models of energy demand or public-health risk assessment. 

The source code for the experiments described in this article can be found in the github repository: \url{https://github.com/EnergyWeatherAI/LST_downscaling_and_nowcasting}. We share the preprocessed machine learning-ready datasets used in this study via Zenodo repository \cite{Kurchaba2026-kg}.

\begin{acks}
This study was conducted as part of the Horizon Europe project UrbanAIR and has received funding from the Swiss State Secretariat for Education, Research and Innovation (SERI)
\end{acks}

\typeout{}


\appendix
\begin{appendices}
\section{Downscaling model specifications}
\label{app:downscaling_specs}
For the downscaling, we use a typical U-Net \cite{ronneberger2015u}, composed of four downsampling (encoder) and upsampling (decoder) stages and skip connections between them. The details of the architecture used can be found in Table \ref{tab:downscaling_archi}. For training, we use the Adam optimizer with a maximum of $1000$ epochs and early stopping set at $10$ epochs for the validation RMSE. The learning rate used is $2 \times 10^{-5}$, and the batch size is $32$. The reproducibility seed is set at $0$. Training time for a model run is approximately $3$ hours on average using one NVIDIA A100 GPU node.

\begin{table*}[t]
\centering
\begin{tabular}{ccccc}
\hline
\textbf{Stage} & \textbf{Operation} & \textbf{Channels} & \textbf{Spatial} & \textbf{Role} \\
\hline

\multicolumn{5}{c}{\textit{Encoder}} \\
\hline
1 & Conv Block & 3 $\rightarrow$ 64 & 128$\times$128 & Feature extraction \\
  & MaxPool (2$\times$2) & 64 $\rightarrow$ 64 & 128$\rightarrow$64 & Downsampling \\
\hline
2 & Conv Block & 64 $\rightarrow$ 128 & 64$\times$64 &  \\
  & MaxPool (2$\times$2) & 128 $\rightarrow$ 128 & 64$\rightarrow$32 & Downsampling \\
\hline
3 & Conv Block & 128 $\rightarrow$ 256 & 32$\times$32 &  \\
  & MaxPool (2$\times$2) & 256 $\rightarrow$ 256 & 32$\rightarrow$16 & Downsampling \\
\hline
4 & Conv Block & 256 $\rightarrow$ 512 & 16$\times$16 &  \\
  & MaxPool (2$\times$2) & 512 $\rightarrow$ 512 & 16$\rightarrow$8 & Downsampling \\
\hline

\multicolumn{5}{c}{\textit{Bottleneck}} \\
\hline
5 & Conv Block & 512 $\rightarrow$ 1024 & 8$\times$8 & Deep representation \\
\hline

\multicolumn{5}{c}{\textit{Decoder}} \\
\hline
6 & Transposed Conv (2$\times$2, s=2) & 1024 $\rightarrow$ 512 & 8$\rightarrow$16 & Upsampling \\
  & Concatenate (skip) & 512 + 512 $\rightarrow$ 1024 & 16$\times$16 & Skip connection \\
  & Conv Block & 1024 $\rightarrow$ 512 & 16$\times$16 & Feature fusion \\
\hline
7 & Transposed Conv (2$\times$2, s=2) & 512 $\rightarrow$ 256 & 16$\rightarrow$32 & Upsampling \\
  & Concatenate (skip) & 256 + 256 $\rightarrow$ 512 & 32$\times$32 & Skip connection \\
  & Conv Block & 512 $\rightarrow$ 256 & 32$\times$32 & Feature fusion \\
\hline
8 & Transposed Conv (2$\times$2, s=2) & 256 $\rightarrow$ 128 & 32$\rightarrow$64 & Upsampling \\
  & Concatenate (skip) & 128 + 128 $\rightarrow$ 256 & 64$\times$64 & Skip connection \\
  & Conv Block & 256 $\rightarrow$ 128 & 64$\times$64 & Feature fusion \\
\hline
9 & Transposed Conv (2$\times$2, s=2) & 128 $\rightarrow$ 64 & 64$\rightarrow$128 & Upsampling \\
  & Concatenate (skip) & 64 + 64 $\rightarrow$ 128 & 128$\times$128 & Skip connection \\
  & Conv Block & 128 $\rightarrow$ 64 & 128$\times$128 & Feature fusion \\
\hline

10 & Conv (1$\times$1) & 64 $\rightarrow$ 1 & 128$\times$128 & Output layer \\
\hline
\end{tabular}
\caption{The downscaling model U-Net architecture used. Convolutional layers are defined by kernel size, stride ($s$), and padding ($p$).}
\label{tab:downscaling_archi}
\end{table*}

\section{Nowcasting model specifications}
\label{app:nowcast_specs}
The nowcasting model architecture is a typical implementation of Convolutional LSTM Network, which is composed of an encoder and a decoder part \cite{shi2015convolutional, app142311315, vukotic2017one}, since the desired output is an image. 
The encoder consists of three hierarchical stages, each combining:
\begin{itemize}
    \item A convolutional block (for feature extraction and downsampling). The kernel size used is $3\times3$.
    \item A ConvLSTM layer (temporal modeling). The kernel size used is $5\times5$.
\end{itemize}
We use the LeakyReLU activation function with a slope set at the value $0.2$. 

The decoder mirrors the encoder in reverse order, combining:
\begin{itemize}
    \item ConvLSTM layers (initialized with encoder outputs),
    \item Transposed convolutions for upsampling with a stride of $2$, kernel size $4 \times 4$, and padding $1$,
    \item Two final convolutional layers. The first convolutional layer uses a $3 \times 3$ kernel and the second a $1 \times 1$ to generate the final $128 \times 128$ prediction.
\end{itemize}
We also use the LeakyReLU activation function with a slope set at the value $0.2$. The architecture is shown in Table~\ref{tab:conv_lstm_archi}. For training, we use the Adam optimizer with a learning rate of $1e-4$, a batch size of $128$, a maximum of $1000$ epochs, and early stopping set at $10$ epochs for the validation RMSE. Training time for a model run is approximately $7$ hours on average for Bucharest using one NVIDIA A100 GPU node. For Berlin and Antwerp, this is reduced to approximately $4$ hours due to less data being available for those cities than for Bucharest. 

\begin{table*}[t]
\centering
\begin{tabular}{ccccc}
\hline
\textbf{Stage} & \textbf{Operation} & \textbf{Channels} & \textbf{Spatial} & \textbf{Role} \\
\hline

\multicolumn{5}{c}{\textit{Encoder}} \\
\hline
\multirow{2}{*}{Stage 1} 
 & Conv (3$\times$3, s=1, p=1) + LeakyReLU & 1 $\rightarrow$ 32 & 128$\times$128 & Feature extraction \\
 & ConvLSTM (5$\times$5) & 32 $\rightarrow$ 64 & 128$\times$128 & Temporal modeling \\
\hline

\multirow{2}{*}{Stage 2} 
 & Conv (3$\times$3, s=2, p=1) + LeakyReLU & 64 $\rightarrow$ 64 & 128$\rightarrow$64 & Downsampling \\
 & ConvLSTM (5$\times$5) & 64 $\rightarrow$ 96 & 64$\times$64 & Temporal modeling \\
\hline

\multirow{2}{*}{Stage 3} 
 & Conv (3$\times$3, s=2, p=1) + LeakyReLU & 96 $\rightarrow$ 96 & 64$\rightarrow$32 & Downsampling \\
 & ConvLSTM (5$\times$5) & 96 $\rightarrow$ 128 & 32$\times$32 & Temporal modeling \\
\hline

\multicolumn{5}{c}{\textit{Decoder}} \\
\hline
\multirow{2}{*}{Stage 3} 
 & ConvLSTM (5$\times$5) & 128 $\rightarrow$ 128 & 32$\times$32 & Temporal modeling \\
 & Transposed Conv (4$\times$4, s=2, p=1) + LeakyReLU & 128 $\rightarrow$ 128 & 32$\rightarrow$64 & Upsampling \\
\hline

\multirow{2}{*}{Stage 2} 
 & ConvLSTM (5$\times$5) & 128 $\rightarrow$ 96 & 64$\times$64 & Temporal modeling \\
 & Transposed Conv (4$\times$4, s=2, p=1) + LeakyReLU & 96 $\rightarrow$ 96 & 64$\rightarrow$128 & Upsampling \\
\hline

\multirow{3}{*}{Stage 1} 
 & ConvLSTM (5$\times$5) & 96 $\rightarrow$ 64 & 128$\times$128 & Temporal modeling \\
 & Conv (3$\times$3, s=1, p=1) + LeakyReLU & 64 $\rightarrow$ 32 & 128$\times$128 & Feature refinement \\
 & Conv (1$\times$1, s=1, p=0) & 32 $\rightarrow$ 1 & 128$\times$128 & Output layer \\
\hline
\end{tabular}
\caption{Architecture of the ConvLSTM-based nowcasting model. Convolutional layers are defined by kernel size, stride ($s$), and padding ($p$).}
\label{tab:conv_lstm_archi}
\end{table*}

\newpage
\section{Results for Antwerp and Berlin}
\label{app:nowcastplotsAntwerpBerlin}

To avoid having too many figures in the main text, we present in this appendix the nowcasting results for Antwerp and Berlin. The conclusions drawn from the results of Bucharest presented in the main text also apply to the figures presented in this section.

\begin{figure*}[b]
\centering

\includegraphics[width=0.45\textwidth]{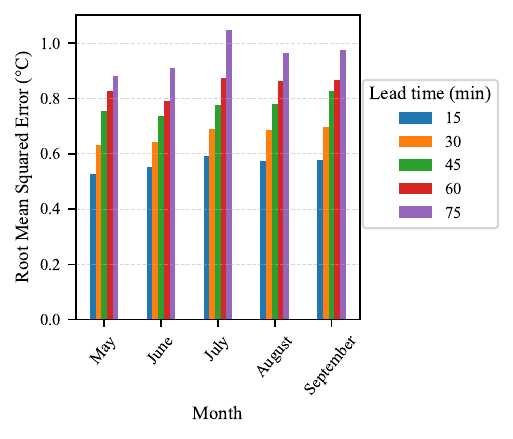}
\hfill
\includegraphics[width=0.45\textwidth]{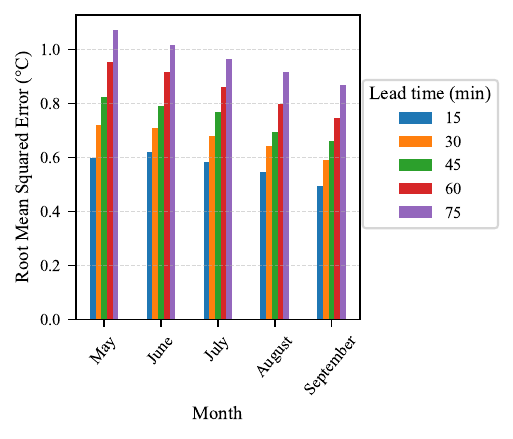}

\vspace{0.5em}

\includegraphics[width=0.45\textwidth]{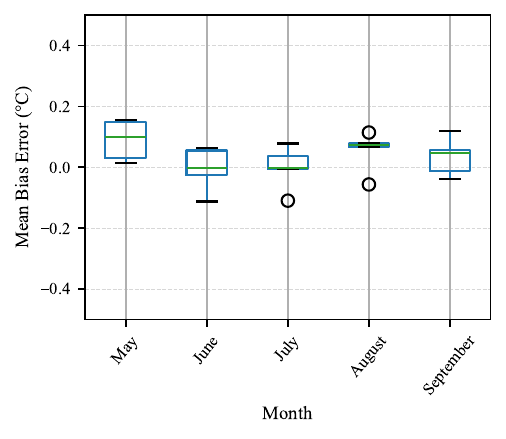}
\hfill
\includegraphics[width=0.45\textwidth]{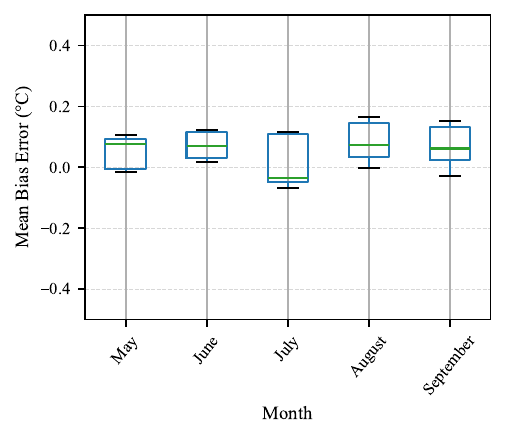}

\caption{
Nowcasting performance per studied month for Antwerp (left) and Berlin (right).
Top row: RMSE per month per lead time.
Bottom row: MBE distribution over lead times.
}
\label{fig:nowcast_antwerp_berlin_all}
\end{figure*}

\clearpage

\begin{figure*}
   \centering
    \includegraphics[width=0.9\textwidth]{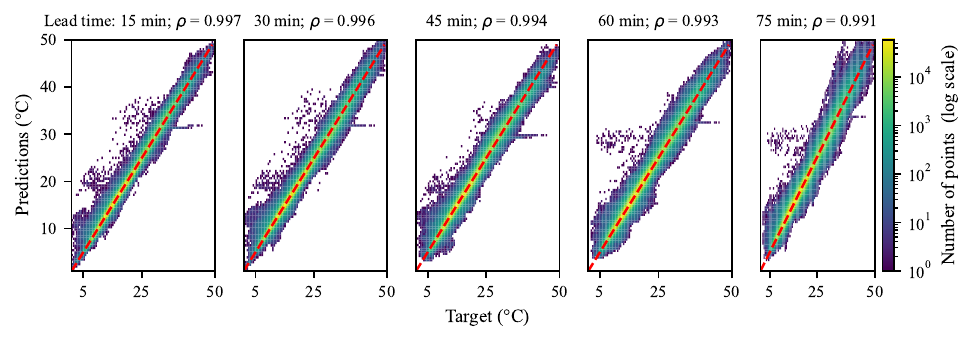}
  \caption{Predictions versus target values of the LST nowcasting model for different lead times for the city of Antwerp. The red line corresponds to 45-degree line. $\rho$ corresponds to the Pearson correlation coefficient.}
    \label{fig:fit_predict_nowcasting_antwerp}
\end{figure*}

\begin{figure*}
   \centering
    \includegraphics[width=1.\textwidth]{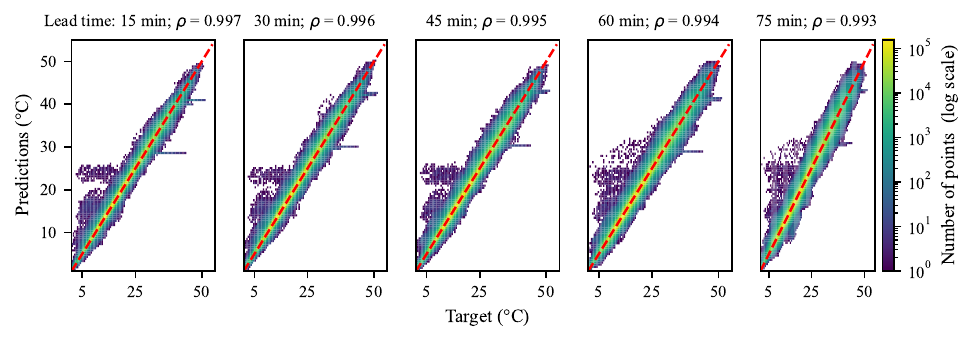}
  \caption{Predictions versus target values of the LST nowcasting model for different lead times for the city of Berlin. The red line corresponds to 45-degree line. $\rho$ corresponds to the Pearson correlation coefficient.}
    \label{fig:fit_predict_nowcasting_berlin}
\end{figure*}

\begin{figure*}
  \centering
  \includegraphics[width=0.9\textwidth]{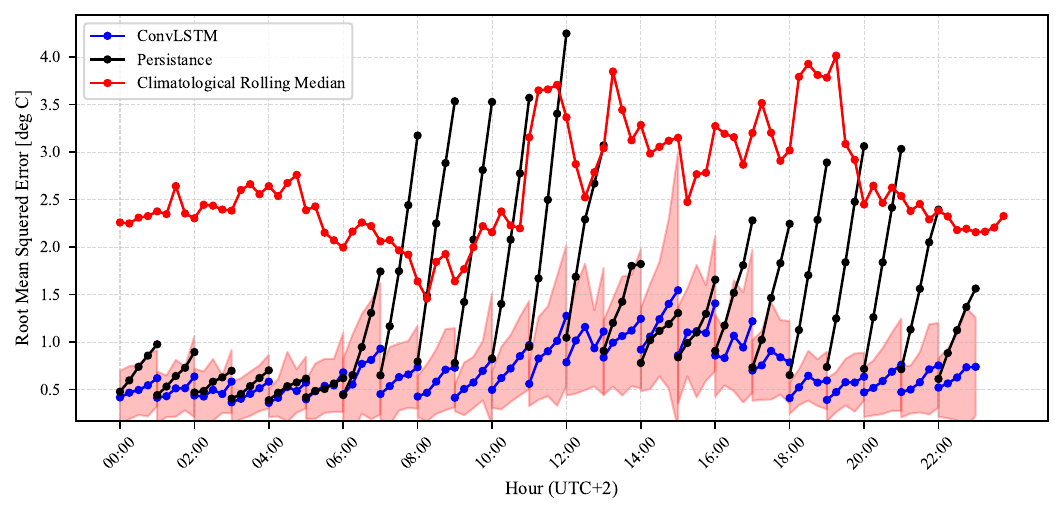}
  \caption{Diurnal evolution of the RMSE per lead time for the city of Antwerp: ConvLSTM-based LST nowcasting model, Persistence benchmark and Climatological Rolling Median benchmark (computed over two consecutive days due to the low data coverage). All times are in local summer time of Antwerp (UTC+2).}
  \label{fig:diurnal_rmse_antwerp}
\end{figure*}

\begin{figure*}
  \centering
  \includegraphics[width=0.9\textwidth]{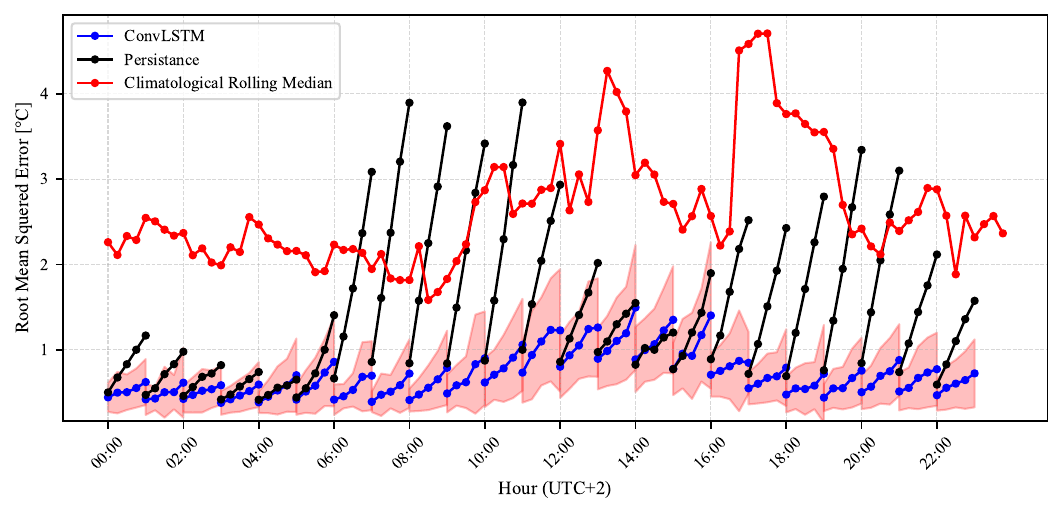}
  \caption{Diurnal evolution of the RMSE per lead time for the city of Berlin: ConvLSTM-based LST nowcasting model, Persistence benchmark and Climatological Rolling Median benchmark (computed over two consecutive days due to the low data coverage). All times are in local summer time of Berlin (UTC+2).}
  \label{fig:diurnal_rmse_berlin}
\end{figure*}

\begin{figure*}
  \centering
  \includegraphics[width=0.8\textwidth]{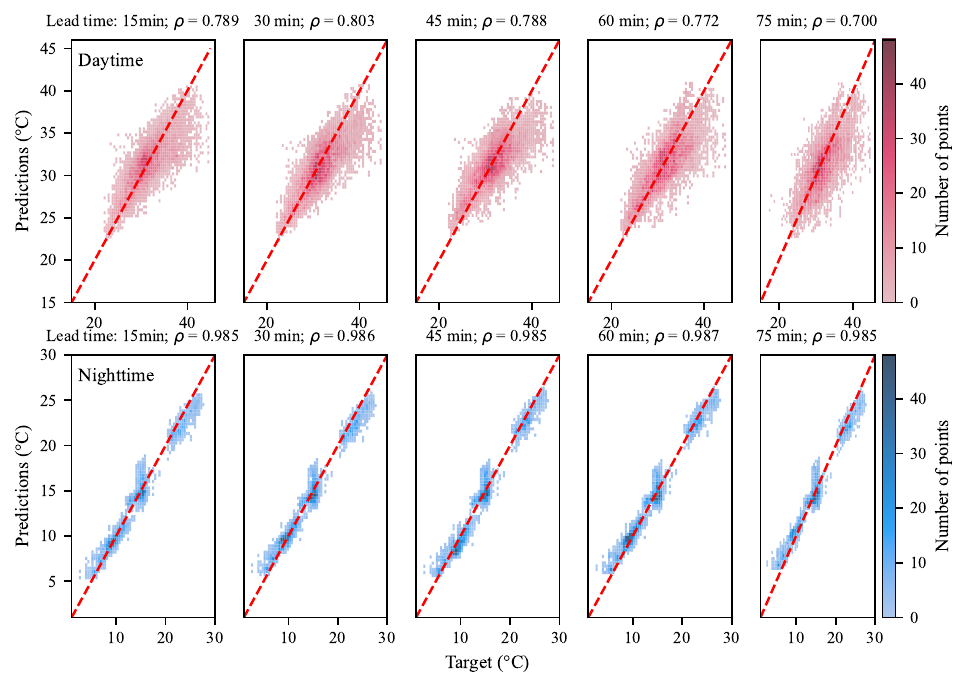}
  \caption{Nowcasting results vs MODIS measurement per lead time for the city of Antwerp. Top panel: LST estimates vs corresponding MODIS-derived LSTs from daytime overpasses of Terra and Aqua. Bottom panel: LST estimates vs corresponding MODIS-derived LSTs from nighttime overpasses of Terra and Aqua. $\rho$ corresponds to the Pearson correlation coefficient.}
  \label{fig:nowcast_vs_modis_antwerp}
\end{figure*}

\begin{figure*}
  \centering
  \includegraphics[width=0.8\textwidth]{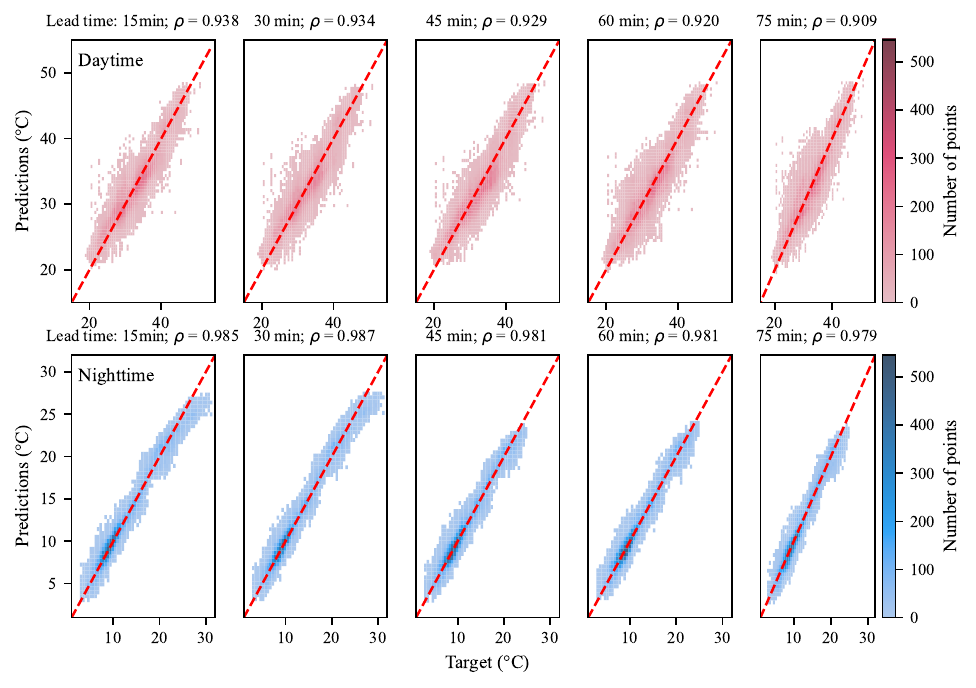}
  \caption{Nowcasting results vs MODIS measurement per lead time for the city of Berlin. Top panel: LST estimates vs corresponding MODIS-derived LSTs from daytime overpasses of Terra and Aqua. Bottom panel: LST estimates vs corresponding MODIS-derived LSTs from nighttime overpasses of Terra and Aqua. $\rho$ corresponds to the Pearson correlation coefficient.}
  \label{fig:nowcast_vs_modis_berlin}
\end{figure*}

\clearpage

\end{appendices}
\bibliographystyle{unsrt}
\bibliography{literature}
\end{document}